\documentclass[a4paper,
               biblatex,     % biblatex is used
               jacowbiblatex,
               ]{jacow}
\usepackage{pdfpages,multirow,ragged2e} %
\makeatletter%
	\ifboolexpr{bool{xetex}}
	 {\renewcommand{\Gin@extensions}{.pdf,%
	                    .png,.jpg,.bmp,.pict,.tif,.psd,.mac,.sga,.tga,.gif,%
	                    .eps,.ps,%
	                    }}{}
\makeatother

\ifboolexpr{bool{xetex} or bool{luatex}} % test for XeTeX/LuaTeX
 {}                                      % input encoding is utf8 by default
 {\usepackage[utf8]{inputenc}}           % switch to utf8

\usepackage[USenglish]{babel}
\usepackage{hyperref}

\newcommand{\code}[1]{\texttt{#1}}
\newcommand{\word}[1]{``\textit{#1}''}
\newcommand{\wv}[1]{\texttt{Word2Vec}}
\newcommand{\simcse}{\texttt{SimCSE}}
\newcommand{\bt}{\texttt{BERTopic}}
\newcommand{\hdbscan}{\texttt{HDBSCAN}}
\newcommand{\lda}{\texttt{LDA}}
\newcommand{\umap}{\texttt{UMAP}}
\newcommand{\etal}{et~al~}
\newcommand{\eg}{e.~g.~}

\newcommand{\fig}[1]{Fig.~\ref{#1}}
\newcommand{\zsim}{\mathbf{z}}

\newcommand{\esp}{\vspace{-0.8em}}
\newcommand{\ssp}{\vspace{-0.75em}}

\ifboolexpr{bool{jacowbiblatex}}%
 {%
  \addbibresource{references.bib}
 }{}
\usepackage{float}
\floatstyle{plaintop}
\restylefloat{table}

\begin{document}

\title{Textual Analysis of ICALEPCS and IPAC Conference Proceedings: Revealing Research Trends, Topics, and Collaborations for Future Insights and Advanced Search}

\author{A. Sulc\thanks{antonin.sulc@desy.de}, A. Eichler, T. Wilksen,  DESY, Hamburg, Germany}
	
\maketitle
\begin{abstract}
In this paper, we show a textual analysis of past ICALEPCS and IPAC conference proceedings to gain insights into the research trends and topics discussed in the field. We use natural language processing techniques to extract meaningful information from the abstracts and papers of past conference proceedings. We extract topics to visualize and identify trends, analyze their evolution to identify emerging research directions, and highlight interesting publications based solely on their content with an analysis of their network. Additionally, we will provide an advanced search tool to better search the existing papers to prevent duplication and easier reference findings. Our analysis provides a comprehensive overview of the research landscape in the field and helps researchers and practitioners to better understand the state-of-the-art and identify areas for future research.
\end{abstract}

\section{Introduction}
The field of language processing has noticed remarkable advances and breakthroughs, shaping our understanding of the fundamental principles of working with written knowledge and our ability to automatically process it.
As research facilities in the field of particle accelerators continue to push the boundaries in improving the particle accelerators and their controls, a large amount of text corpus has been created, capturing the collective knowledge and discoveries in the community.
In this paper, we explore the rich archive of past papers on the particle accelerator from IPAC and ICALEPCS conference proceedings, employing recent language processing techniques.
We aim to expose the evolution of ideas, explore the interconnectedness of research areas, and provide a tool for advanced search and discovery that adapts to the language used in the community.

Faced with many potentially relevant papers, researchers must either narrowly limit the scope of their review or rely on new methods to efficiently analyze large document collections.
One of the key goals of this work is to help research orient in the overwhelmingly many papers being introduced every year in the community of particle accelerators.
To overcome these limitations, researchers are increasingly adopting automated methods for topic modeling~\cite{asmussen2019smart}, semantic search, and knowledge extraction that can rapidly analyze word patterns across thousands of documents to reveal latent thematic structures.

For example, as researchers disseminate their findings, they reference and extend prior work, creating connections within a knowledge network. A published study may later be cited by another, building links that indicate the flow of knowledge. Highly cited papers and their authors often represent influential roles within this network. There are also links between individual papers based solely on their text content, which, exposed to other papers, creates a hidden underlying complex network that can reveal papers that for instance cover important topics.
Additionally, papers can be characterized by the topics they address, and their evoluation. Each paper encapsulates a specific topic that may be relevant to a particular sub-community or provide insights into the development of a specific field.
Moreover, semantic search plays a pivotal role in unveiling these concealed thematic structures. It empowers scientists to explore textual data beyond traditional full-text search, mitigating issues like typographical errors and synonyms.

This paper serves as an exploration into the possibilities of automatically exposing the collective wisdom stored within the particle accelerator community through contributions presented at the IPAC and ICALEPCS conferences. It provides a unique perspective on the historical evolution of this field, shedding light on previously unnoticed details and noteworthy contributions that might otherwise go unnoticed and help scientists find potential links between their work and potential future collaborations.

The structure of this paper is as follows:
First, we provide background on existing approaches for topic modeling
of complex corpora, summarizing the current landscape of automated paper processing.
We then describe our methodology for enabling robust semantic search across the conference proceedings. This capability serves as the foundation for the analyses that follow, while also providing an invaluable tool for community members to efficiently find relevant papers.
Building on this semantic search, we detail our techniques for uncovering latent topics in the corpus and analyzing their evolution over time for both the IPAC and ICALEPCS conferences.
Finally, we introduce our approach to extracting knowledge from the abstracts using social network analysis of citation patterns. This reveals influential contributions and concepts based on both textual content and the networks formed by citations between papers.
\ssp
\section{Related Work}
\paragraph{Topic Modeling}
Blei~\etal{}~\cite{blei2003latent} pioneered unsupervised topic modeling with Latent Dirichlet Allocation (\lda{}).
\lda{} represents documents as mixtures of topics, where a topic is a distribution over words.
Using Bayesian inference, \lda{} reverse engineers the latent per-topic word distributions and per-document topic mixtures, enabling unsupervised extraction of coherent topics and quantification of document similarities.
With the increasing popularity of dense vector representation~\cite{gao2021simcse,reimers2019sentencebert}~Grootendorst~\cite{grootendorst2022bertopic} introduced \bt{}.
It is an unsupervised topic modeling technique that leverages the vector representation capabilities of modern embedding techniques like~\cite{gao2021simcse,reimers2019sentencebert} to create representations of topics that are more coherent and interpretable compared to traditional topic modeling approaches like \cite{blei2003latent}.
In the domain of graph knowledge representation with neural topic modeling, a notable work of~\cite{zhou2020neural} proposes the Graph Topic Model.
Their approach is based on graph neural network-based topic modeling that represents a text corpus as a graph with documents and words as nodes, uses edges to capture co-occurrence relationships, and leverages graph convolutions to aggregate topical information from neighboring nodes.
From their approach, we borrow the bipartite graph modeling with document-word pairs.

Analysis of textual data from a large text corpus is not new. It has been successfully applied mostly in journalism~\cite{jacobi2018quantitative,dimaggio2013exploiting}.
In~\cite{dimaggio2013exploiting} they analyzed a structure of collections of newspapers. They showed the value of topic modeling with an automated \lda{} analysis for funding applications. Similarly, in~\cite{jacobi2018quantitative} they performed a thorough analysis of the possibilities of applications of topic modeling in journalism.
More recently, the work of Asmussen~\etal{}~\cite{asmussen2019smart} presented a use-case of topic modeling on a large collection of papers for an exploratory literature review.

\paragraph{Semantic Search}
Traditional keyword-based search often relies on simply matching input query words to text. This can be very limiting, as it is sensitive to spelling and typographic errors and neglects the context of the searched words. Approximate string matching techniques like Levenshtein distance can minimize issues caused by minor errors but still ignore semantic context, like synonyms. Furthermore, running Levenshtein distance on a large corpus is computationally very expensive.

Notable techniques that consider word context include \wv{} by Mikolov~\etal{}~\cite{mikolov2013distributed,mikolov2013efficient} and \simcse{}~by Gao~\etal{}~\cite{gao2021simcse}. \wv{} pioneered embedding words in vector spaces where semantically similar words are close together. As a self-supervised autoencoder, it learns to embed words based on nearby context. It embeds individual words but neglects sentence meaning. There are techniques to approximate sentence embeddings with \wv{}.  One approach is summing up the embedding of every word in the sentence but this approach becomes particularly inaccurate for longer blocks of text because summation ignores the position of words in the sentence.
Recent self-supervised methods, like by \simcse{}, have effectively tackled this challenge by embedding entire sentences and training models to understand sentence similarities through contrastive learning losses. At the heart of \simcse{} lies the concept of sentence embedding, achieved by generating contrastive samples from pairs of identical text segments with slight modifications induced by standard dropout techniques. This ingenious approach allows the model to semantic understanding from unlabeled data without the need for annotations, thus capturing meaning without relying on predefined labels.
Notably, \simcse{} also incorporates advanced tokenization methods, which serve to mitigate the impact of writing errors. In cases where unknown tokens arise, these tokens undergo a splitting process and can still be embedded like error-free tokens, ensuring robustness in the face of textual imperfections.

By taking context into account, techniques like \wv{} and \simcse{} transcend the limitations of traditional keyword-based representations, addressing issues such as synonyms and broadening the horizons of text discovery, and both \wv{} and \simcse{} play an important role in embedding abstracts for topic modeling, serving as essential tools not only for efficient searching but also for meaningful content interpretation.

\ssp
\section{Advanced Text Search and Embedding}
In this section, we detail our approach for representing academic abstracts as dense vector embeddings using \simcse{} and \wv{}.
\ssp
\subsection{\simcse}
To enable semantic search, we embed abstracts into vector representations using \simcse{}, a semantic textual similarity model. While \simcse{} provides an accurate pre-trained model, its source corpus performs poorly on domain-specific problems like ours. Therefore, we fine-tuned \simcse{} on our full corpus of scientific papers to adapt it to our task. \fig{fig:simcse} showing improved performance of our fine-tuned model compared to the original pre-trained \simcse{}.

For each abstract, \simcse{} encodes the text into an $M$-dimensional ($M=768$ in our case which is imposed by the dimension of a pre-trained model) vector $\zsim{}\in \mathbb{R}^{M}$ that captures semantic content and context.

To enable efficient similarity comparisons, we normalize $\zsim{}$ to unit length and denote it as $\hat{\zsim{}}$ ($\|\hat{\zsim{}}\|_2=1$).
This allows us to compute relevance between a query vector $\hat{\zsim{}}_q$ and all abstract embeddings $\hat{\zsim{}}_i$ via a simple dot product, without needing to re-normalize all vectors embeddings for each query.
The dot product yields cosine similarities in the range $[-1, 1]$, with 1 indicating maximum similarity and -1 indicating maximum dissimilarity.
\ssp
\subsection{\wv{}}
While \simcse{} shows impressive performance in encoding sentences into dense vectors, it may not be optimal if one wants to look for keywords instead of sentences, because \simcse{} is designed to handle entire sentences.
As an alternative to \simcse{} vector representations, we use \wv{} embedding for matching individual keywords.
This approach matches the individual token embeddings for each query keyword (or near variants based on Levenshtein distance) against all tokens in the abstracts using trained \wv{} embeddings.
Important is that \wv{} search can be beneficial for queries with just a few keywords, where each word may have multiple meanings (\eg{}~ \code{barrier}, \code{film}, and \code{membrane} are synonyms).
In \wv{} search, we represent each keyword and abstract token as a set of vectors based on the \wv{} embeddings. Matching is done by computing cosine similarity between the query vectors and every word vector in each abstract.
The final ranking is done by sorting based on the product of cosine similarities across all query vectors for each abstract. \wv{} search provides a flexible keyword-based alternative to \simcse{} for identifying relevant abstracts based on the semantic similarity of individual words.
\ssp
\section{Topic Modeling}
Automated topic modeling like \lda{}~\cite{blei2003latent} and \bt{}~\cite{grootendorst2022bertopic} are powerful unsupervised methods for uncovering latent semantic structures within large corpora.
By applying topic modeling to archives of conference abstracts, we can gain insights into the evolution of ideas and priorities within a research community over time.
In this section, we show some results about the development of the accelerator science field by mining collections of conference abstracts from IPAC (2010-2023) and ICALEPCS (1999-2021) using the \bt{} framework.
We first extract and transform abstract texts into a suitable format, then perform data analysis to identify semantic topics and their connections. One could also consider incorporating entire papers, but since we are interested in mostly the landscape of the domain, abstracts are sufficient for our purpose.

We use \bt{} as it serves better for our purpose. Apart from many conceptual differences between \lda{} and \bt{}, \lda{} assigns multiple topics per document, while \bt{} assigns a single topic to each document (abstract).
This is an important distinction for our goal of mapping the conceptual landscape, as single topic assignment provides clearer interpretability.
Additionally, \bt{} allow custom embeddings, which can be advantageous in a domain-specific scenario, with our fine-tuned \simcse{} embedding.

To summarize, the \bt{} algorithm performs the following key steps:
First, the input documents (abstracts) are embedded into dense vector representations using pre-trained contextual embeddings like \simcse{}. In this step, we leverage our own fine-tuned \simcse{} model to encode each document (abstract).
The dimension $M$ of these semantic vectors tends to have a very high dimension which is impractical (768 in our case), so dimensionality reduction is applied using \umap{}~\cite{mcinnes2018umap} to compress the vectors while retaining structure.
The reduced vectors are then hierarchically clustered using \hdbscan, revealing relationships between discrete topic clusters.
Finally, important terms are extracted from each cluster using class-based TF-IDF to generate interpretable topic labels.
\ssp
\section{Knowledge Extraction}
Graphs offer an efficient method for representing the intricate associations among entities and concepts present in textual information. Moreover, adopting a graph-based structure to depict knowledge enables the representation of intricate relationships and contextual connections. This approach provides supplementary insights, ultimately unveiling connections between entities.

Within this segment, we present two analyses. The first analysis constructs a graph based on a citation network. This approach can potentially uncover significant works to discover, offering hints for exploring worthy sources, and what other works are citing.

The second analysis creates a document-word bipartite graph.
This graph involves two types of nodes: one for documents and the other for unique words. Connections are established whenever a word appears in the abstract of a document, creating links where direct associations between documents and words are present. Subsequently, we introduce a collection of operations tailored to this type of graph based on bipartite graph projection and several graph metrics, see \fig{fig:bipartitegraph}.
\ssp
\subsection{Citation Graph}
First, we extracted citations from papers, matched them with the available papers, and created a directed graph of references. This builds a very sparse graph.
To measure the importance of a document (node in the graph), there are several centrality metrics, which can assign importance to the node based on connections (citations).
The most straightforward centrality is closeness centrality, which measures the average shortest path length from a node to all other nodes, identifying nodes that can most efficiently spread information. Betweenness centrality counts the number of shortest paths passing through each node, indicating nodes with higher overall influence on network flow. Eigenvector centrality assigns values based on a node's connections to other high-scoring nodes, identifying well-connected prestigious nodes.
Furthermore, one can perform a common neighbors link prediction on individual papers, to reveal potentially hidden connections between adjacent papers that might be interesting to read: when two nodes (documents) have multiple common neighbors, it means that these neighbors are likely to have something in common and might be worth to read (citation).

\begin{figure}[!ht]
    \centering
    \begin{tabular}{c}
    \includegraphics[width=1.0\linewidth]{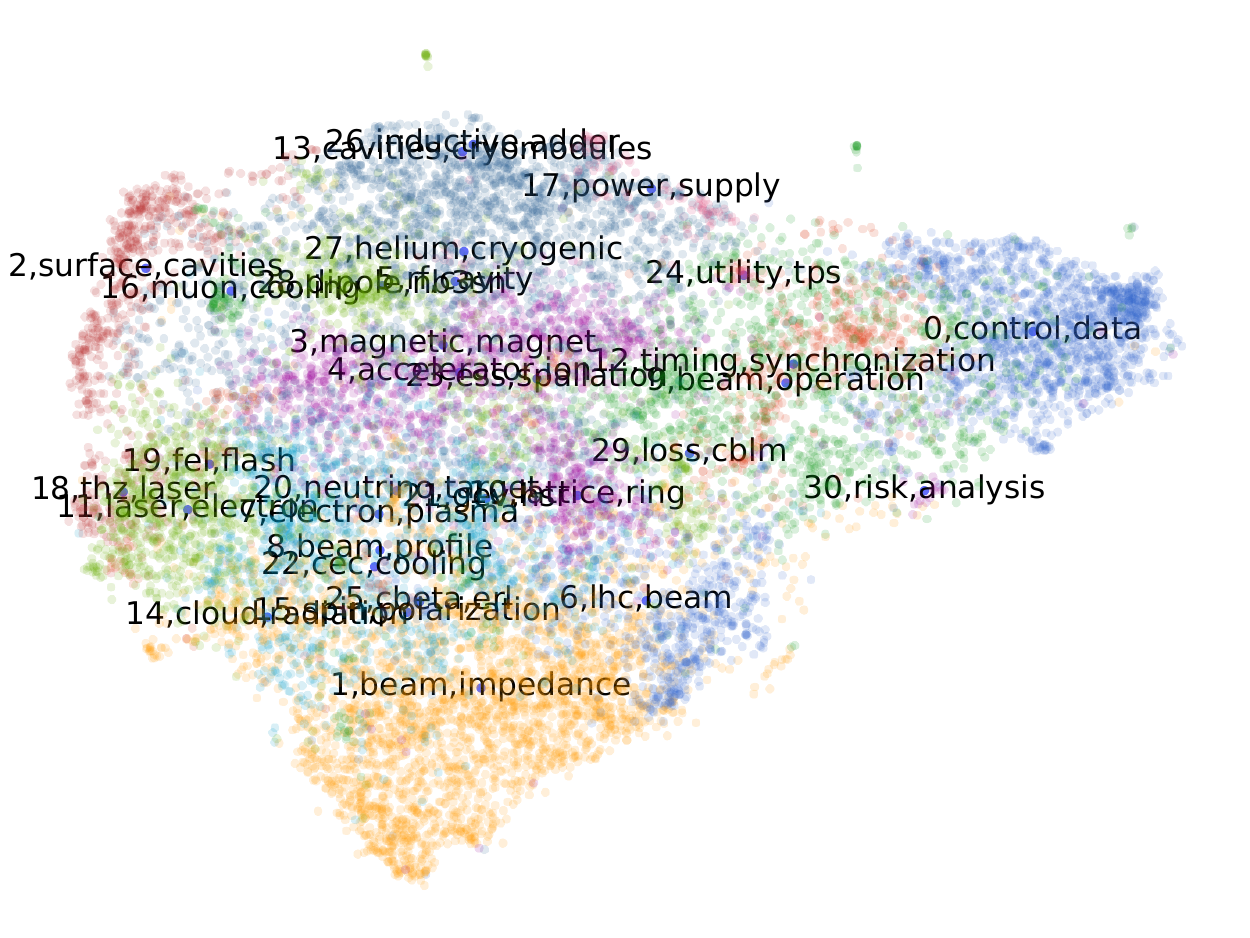}\\
    \includegraphics[width=1.0\linewidth]{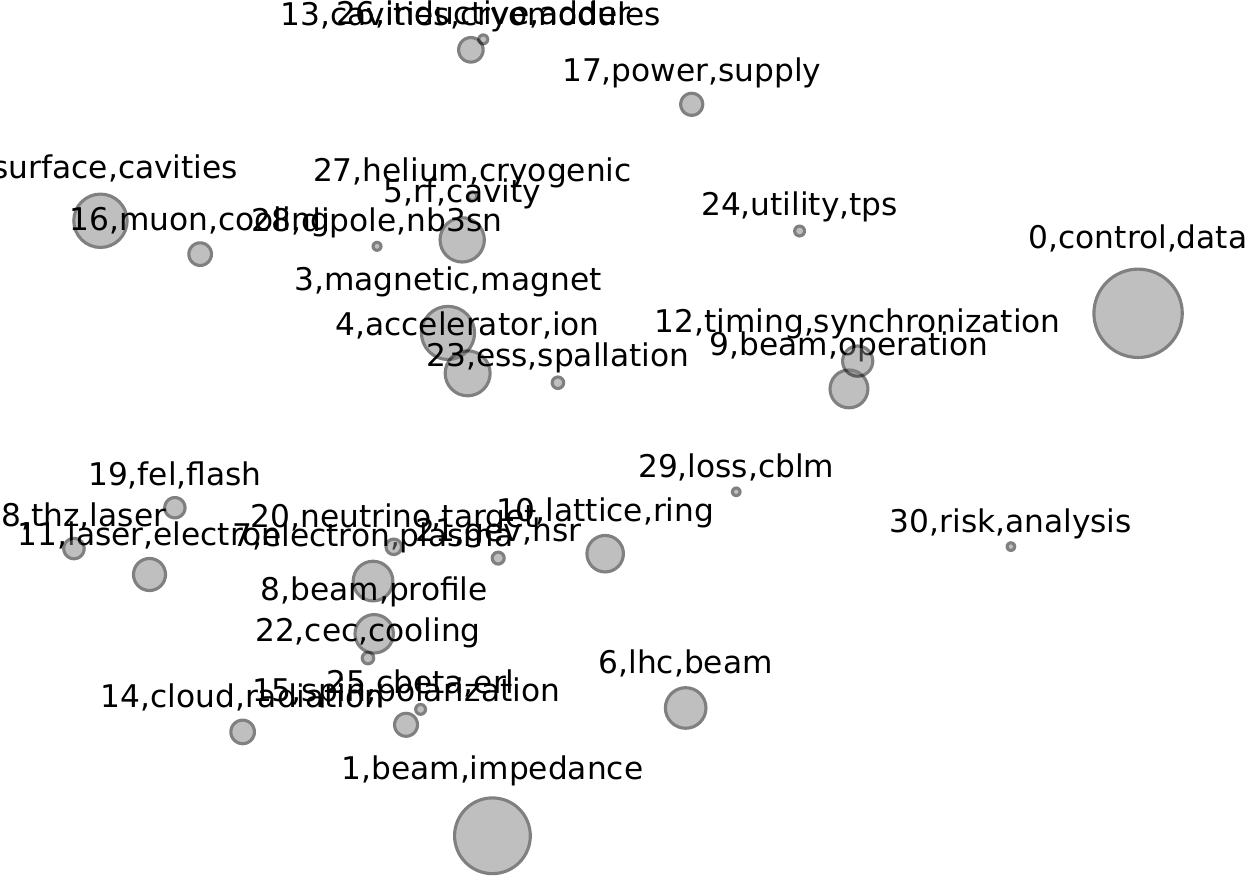}    
    \end{tabular}
    \caption{The upper image shows 32 color-coded topics found with \bt{}. Each point signifies an individual document and is color-coded according to its associated topic.
    The central area of each document embedding shows the topic's keywords.
    In the lower image, we see a two-dimensional distribution of embedded topics. Each circle corresponds to a topic, sized proportionally to its representation within the corpus (the number of papers).
    The position of each topic is determined by the average \umap{} embedding of all documents within that particular topic.
    }
    \label{fig:bertopic}
\end{figure}

\begin{figure}[!ht]
    \centering
    \begin{tabular}{c}
    IPAC \\
    \includegraphics[width=1.0\linewidth]{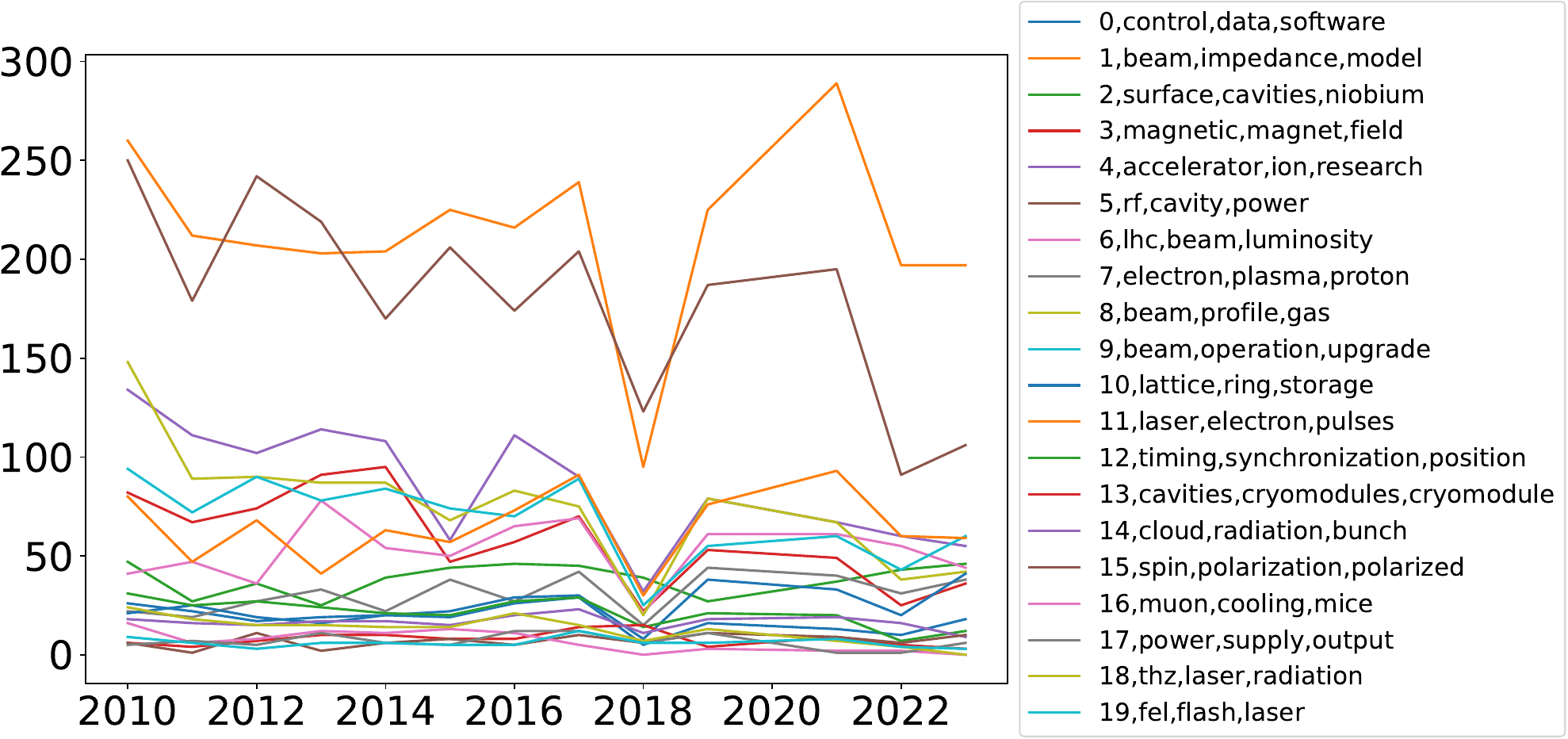}\\
    ICALEPCS\\
    \includegraphics[width=1.0\linewidth]{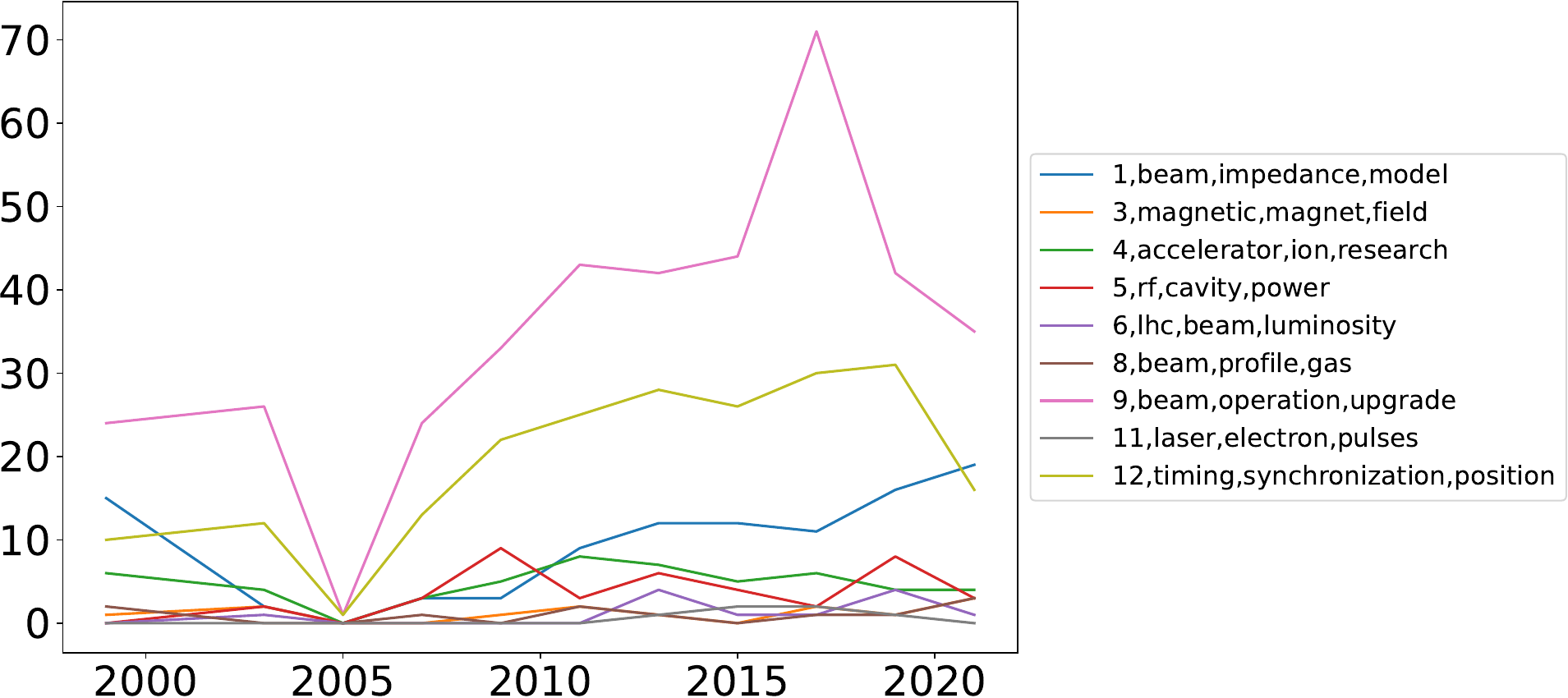}
    \end{tabular}
    \caption{Evaluation of the temporal evolution of the most prominent topics within IPAC (top) and ICALEPCS (bottom) conferences. The data from IPAC 2020 is omitted due to the conference's unusually low number of submissions.
    Furthermore, topic 0 (controls) is concealed in the ICALEPCS analysis, as it is disproportionately over-represented.}
    \label{fig:bertopictime}
\end{figure}

\ssp
\subsection{Document-Word Bipartite Graph}
\begin{figure}[ht!]
    \centering
    \includegraphics{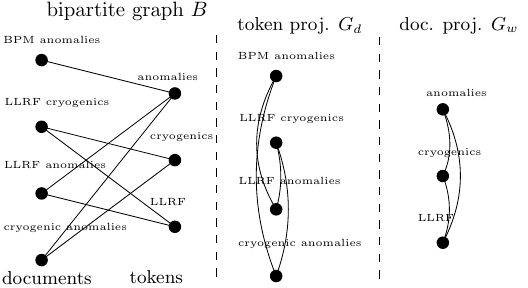}
    \caption{
    The figure shows an example bipartite graph B with two node sets - 4 documents on the left (\code{BPM anomalies}, \code{LLRF cryogenics}, \code{LLRF anomalies} and \code{cryogenics anomalies}) and unique words on the right.
    Edges connect each document node to the word nodes contained in that document. For example, \code{LLRF anomalies} link to \code{LLRF} and \code{anomalies}.
    Projecting the word nodes creates a document graph, connecting documents that share these words. \code{LLRF anomalies} and \code{cryogenics anomalies} link since they both contain \code{anomalies}.
    Similarly, projecting the document nodes creates a word graph, linking words that appear together. \code{LLRF} and \code{cryogenics} link since they co-occur in \code{LLRF cryogenics}.
    The projections transform the original bipartite graph into interconnected documents and word networks.}
    \label{fig:bipartitegraph}
\end{figure}

We used a bipartite knowledge extraction approach inspired by~\cite{zhou2020neural}, see \fig{fig:bipartitegraph}.

A bipartite graph $B$ consists of two disjoint node sets, where edges only connect nodes from one set to the other. There are no edges between nodes within the same set.
In our case, the two node types are documents (abstracts) and unique terms. Edges connect documents to the terms (tokens) they contain.

An important bipartite graph operation is projection - forming a new graph using nodes from one set, with edges for nodes sharing a neighbor in the original graph.
Projecting the document and term nodes creates two new graphs, revealing insights about the corpus: 1) Projecting terms connect documents with similar words in a graph $G_d$, enabling document centrality analysis to identify significantly interconnected papers. 2) Projecting documents link common words in a graph $G_w$, highlighting key terminology based on interconnectedness.

Bipartite graph projection distills the corpus into an interconnected document and term networks. Analyzing these graphs exposes influential contributions and concepts in the community, however, it is important to bear in mind that projections of the bipartite graph $B$ don't indicate the explicit popularity of particular papers or tokens, which is better covered with citation graphs. Instead, it unveils the papers with substantial coverage, either through the highest number of mentions in the case of $G_w$, or the words covered in the case of $G_d$.
\ssp
\section{Results}
\ssp
\subsubsection{Topic Modeling}
\begin{figure*}
    \centering
    \resizebox{0.9\linewidth}{!}{
    \begin{tabular}{cc}
		 \includegraphics[width=0.5\linewidth]{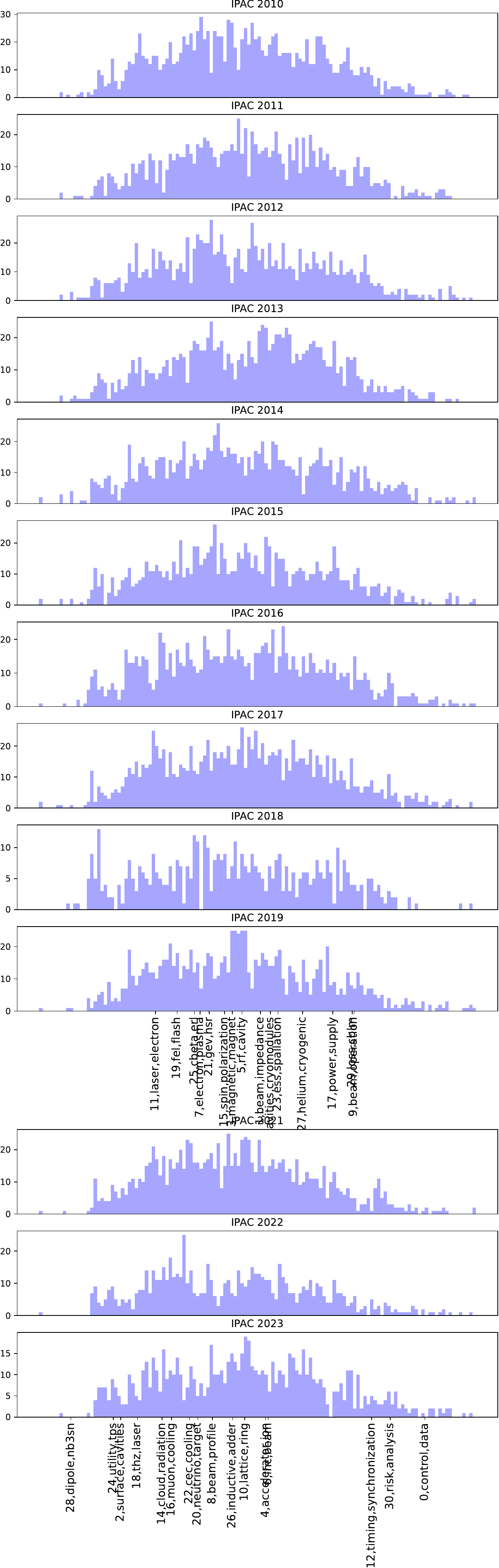} & \includegraphics[width=0.5\linewidth]{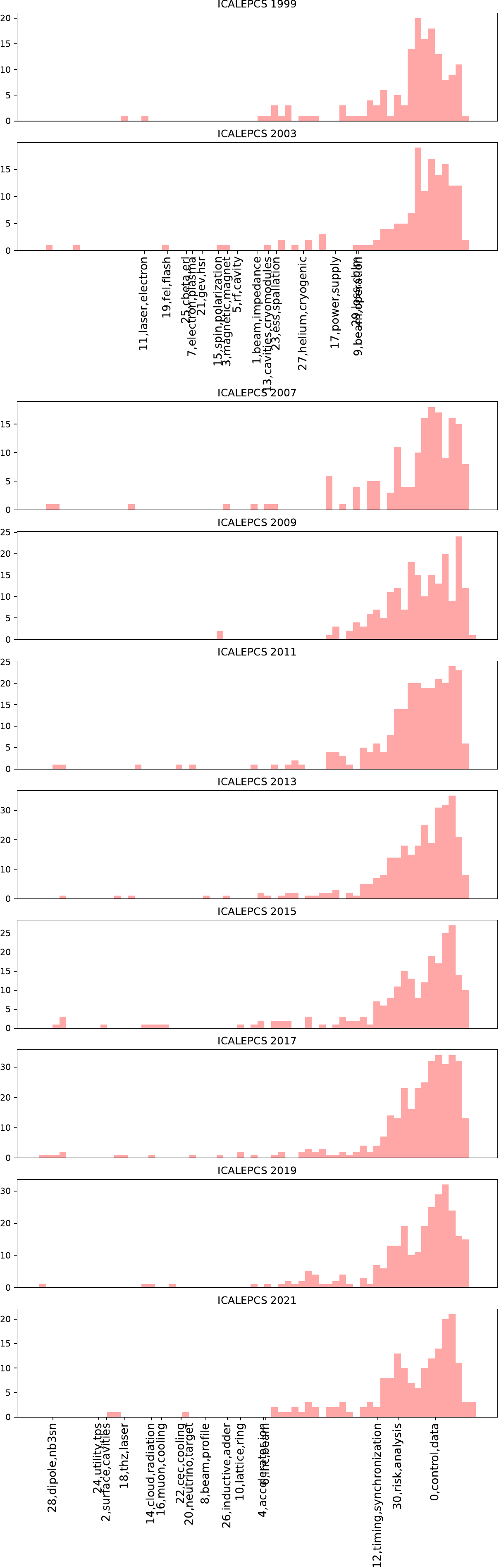}  
    \end{tabular}}
    \caption{Distributions of papers binned (128 bins IPAC and 64 bins for ICALEPCS) after their embeddings are projected into 1D space with \umap{}. The below graphs show likely topic locations (mean of the embeddings). Visualization shows the evolution of topics over time by changing the volume of the papers over the years. }
    \label{fig:BERTvolume}
\end{figure*}

The topic modeling results reveal several insightful trends about the evolution of accelerator and controls research, see \fig{fig:bertopic}, \fig{fig:bertopictime}, and \fig{fig:BERTvolume}.

Most notably, we find that control research (topic 0) forms a distinct cluster, separate from other IPAC and ICALEPCS topics. This likely reflects the large proportion of controls-focused research presented at ICALEPCS conferences.

Additionally, we identify a topic cluster related to superconducting cavities (topics 2, 5, 13) and associated cryogenics (topic 27) and RF power supply systems (topic 17).

Examining topic trends over time on ICALEPCS publications yields another interesting finding. Beam analysis research (topic 1) gradually gained prominence over the years, while beam operations (topic 9) and timing synchronization work (topic 12) declined in relative share. This points to a shift in control priorities toward beam and machine.

The results presented in \ref{fig:bertopictime} have one notable limitation: each topic is represented by a single category, which fails to show the potentially informative distribution of topics that could be better visualized in~\fig{fig:BERTvolume}.
A closer look at~\fig{fig:BERTvolume} reveals that there is a growing trend in ICALEPCS in topics that are close to control and risk and analysis (topic 30) in recent years. This is likely due to a growing number of machine learning-related submissions.
A similar trend is visible in IPAC, where growth between control (topics 0) and risk and analysis (topic 30) is especially noticeable in IPAC 2021.  
In IPAC 2019 there is a visible increase in the cavity (topic 5) and magnet-related (topic 3) research, this can be caused by the commissioning of some facilities.

Some abbreviations appear in topic keywords, like TPS (Taiwan Photon Source), CBLM (Cerenkov Beam Loss Monitor), Nb3Sn (niobium-tin), and CEC (Coherent Electron Cooling).

\ssp
\subsubsection{Citation Graph}

The three papers with the highest closeness centrality are
\begin{enumerate}
	\esp{}
	\item W. Decking \etal{}, ``Status of the European XFEL'', IPAC'13
	\esp{}
	\item D. Noelle, ``Status of the Standard Diagnostic Systems of the European XFEL'', IBIC'14
	\esp{}
	\item B. Keil et al., ``The European XFEL Beam Position Monitor System'', IPAC'10
\end{enumerate}
\esp{}
betweenness centrality,
\begin{enumerate}
	\esp{}
	\item W. Decking \etal{}, ``Status of the European XFEL'', IPAC'13
	\esp{}
	\item A. Grudiev and W. Wuensch, ``A New Local Field Quantity Describing the High Gradient Limit of Accelerating Structures'', LINAC'08
	\esp{}
	\item K. Ko and A. E. Candel, ``Advances in Parallel Electromagnetic Codes for Accelerator Science and Development'', LINAC'10
\end{enumerate}
\esp{}
eigenvector centrality,
\begin{enumerate}
    \esp{}
	\item A. P. Letchford \etal, ``Status of the RAL Front End Test Stand'', EPAC'15
	\esp{}
	\item M. A. Clarke-Gayther, ``A Fast Beam Chopper for Next Generation High Power Proton Drivers'', EPAC'04
	\esp{}
	\item A. P. Letchford et al., ``Status of the FETS Project'', LINAC'14
	\esp{}
\end{enumerate}

It is important to stress that the graph is very sparsely connected and it is likely that some links were missed due to errors during the extractions.
Additionally, some paper titles appeared in different conferences and although their reference was counted, it might be linked to a wrong paper.
\ssp
\subsubsection{Document-Word Bipartite Graph}
From the initial document-term bipartite graph $B$, two new graphs were constructed. One, denoted as $G_d$, resulted from projecting terms and contains a graph of documents related by word co-occurrence. The other, termed $G_w$, was obtained by projecting documents and forming a term graph connected by shared documents.

In $G_w$, nodes with the highest eigenvector centrality values are: \code{beam} (0.14), \code{bunch} (0.11), \code{cavity} (0.1), \code{rf} (0.1), \code{electron} (0.09), \code{linac} (0.09), \code{lhc} (0.08), \code{fel} (0.08), \code{cavities} (0.08), \code{laser} (0.08). A similar trend appears for degree centrality (edge count): \code{beam} (0.07), \code{cavity} (0.05), \code{bunch} (0.05), \code{rf} (0.05), \code{control} (0.04), \code{lhc} (0.04), \code{laser} (0.04), \code{cavities} (0.04), \code{fel} (0.04), \code{electron} (0.04). These outcomes suggest that research concerning beams/bunches holds substantial influence within the community, as does cavity-related research and development.

Notably, interesting findings emerge from the centralities of $G_w$. A noticeable gap is observed in eigenvector centrality, absent in degree centrality. This suggests the existence of two distinct work communities, this requires closer inspection since no obvious justification was found.
Three papers with the highest degree of centrality are
\begin{enumerate}
	\esp{}
	\item L.~Medina~\etal{}, ``Cavity Control Modelling for SPS-to-LHC Beam Transfer Studies'', IPAC'21
	\esp{}
	\item M.~Chung~\etal{}, ``Transient Beam Loading Effects in Gas-filled RF Cavities for a Muon Collider'', IPAC'13
	\esp{}
	\item K.~Yonehara~\etal{}, ``''R\&D of a Gas-Filled RF Beam Profile Monitor for Intense Neutrino Beam Experiments'', IPAC'17
\end{enumerate}
\esp{}
Similar results for eigenvector centralities, except the third paper titled "Simulations of Head-on Beam-Beam Compensation at RHIC and LHC" presented at IPAC'10.
It is important to stress that these results, unlike the citation graph, do not refer to the paper's popularity, but rather measure the paper's coverage of all words inside the community.

\ssp
\section{Technical Details}
\ssp
\subsection{Obtaining the Corpus and Data Extraction}
We developed a tool to navigate conference websites like JACoW~\cite{jacow2023jacow} and retrieve papers as PDF files, compiling a corpus with each file representing one paper. Due to inconsistent formatting, we had to skip some papers.

The extraction text from the PDFs was done with \code{Pymupdf}~\cite{mckie2023pymupdf}  package, which automatically handles complex layouts like two-column formatting. \code{Pymupdf} outputs text blocks with coordinates and raw text.
From the produced texts, we extracted titles by taking the first all-uppercase text block.
For abstracts, we concatenated all text blocks, then extracted the text between \code{Abstract} and \code{INTRODUCTION} using the regular expression \code{Abstract(.*?)(?:INTRODUCTION)}. In case that tokenized abstract contains less than 50\% of English words~\cite{dwyl2019dict}, it is ignored. This happens when PDFs can't be decoded.

We operated with two datasets, when it is not stated otherwise, we use only IPAC (years between 2010 and 2023) and ICALEPCS (years between 1999 and 2021) proceedings, but for some tasks, it is beneficial to have more data available, so we processed papers from following conferences~\cite{jacow2023jacow}: BIW (years between 2008-2012), CYCLOTRONS (years between 2001 - 2019),  DIPAC (years between 1999 - 2011),  EPAC (years between 1996 - 2008), FEL (years between 2004 - 2019), IBIC (years between 2012 - 2022),  LINAC (years between 1996 - 2022), NAPAC (years between 2011 and 2022), PCAPAC (years between 2008 - 2022), SRF (years between 1999 - 2021). We refer to the dataset with all papers as \code{larger} dataset.
\ssp
\subsection{Fine-tuning Embedding Models}

\begin{figure}
    \centering
    \includegraphics[width=1.0\linewidth]{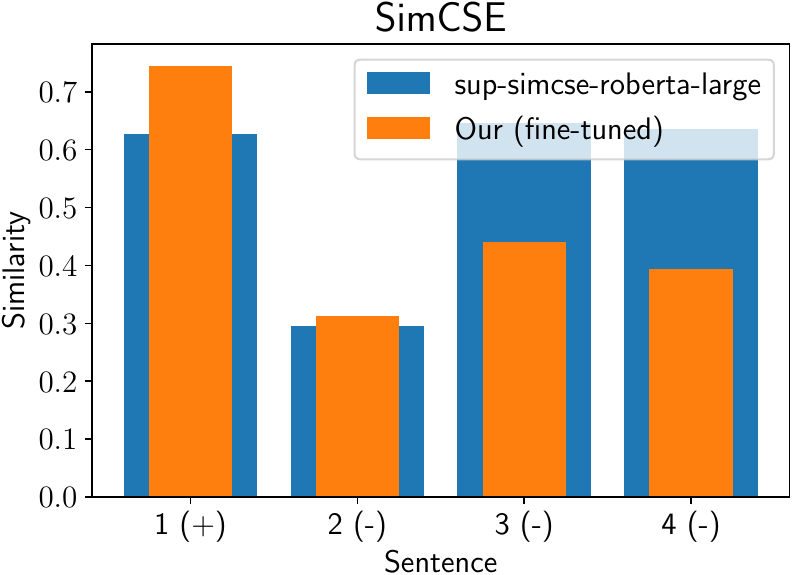}
    \caption{An example demonstrating why our domain-specific fine-tuning is beneficial: The bar chart compares similarity scores (cosine similarity) on intentionally confusing sentences. The query sentence is \word{DESY radio frequency cavities detuned}, compared with: (1) \word{XFEL cavities detuned}, (2) \word{My teeth have frequent cavities}, (3) \word{Please tune radio at low frequency}, and (4) \word{DESY is following a radio at low volume}.
    We compared our fine-tuned model against the state-of-the-art SimCSE model by Gao et al.~\cite{gao2021simcse} (\code{sup-simcse-roberta-large}).
    The first case is an obvious match and both models recognize this, as with the second out-of-context case. However, the third and fourth sentences are correctly identified as rather dissimilar by our model, unlike the pre-trained baseline.
    }
    \label{fig:simcse}
\end{figure}

\subsubsection{\simcse{}}
\label{sect:details:simcse}
For embedding and search, we fine-tuned \simcse{}~\cite{gao2021simcse} in an unsupervised way on the \code{larger} dataset.  %
Although the models provided by~\cite{gao2021simcse} demonstrate a capacity to deal with challenging tasks with tricky grammar, the model is not pre-trained to some peculiar cases specific to our domain as~\fig{fig:simcse} justifies.

The core of the unsupervised training of \simcse{} is in predicting the input sentence itself with only standard dropout.

We fine-tuned \simcse{} using the \code{roberta-base} model and mean pooling of the final layer representations, following the original implementation. To reduce computational demands, we limited the maximum number of tokens per sentence to 256 during training. This is lower than typical values but was necessary to accommodate the large batch size of 128, which provides sufficient negative examples within each batch for contrastive learning.
We used AdamW optimizer with learning rate $\lambda=10^{-5}$. To stabilize the initial convergence of AdamW, we took an initial 50\% batches of the first epoch as warm-up iterations.
We trained the model on the \code{larger} dataset.
\ssp
\subsection{Topic Modeling}
For embedding abstracts, we used our fine-tuned \simcse{} embeddings.
We experimented with a different number of topics, 32 topics turned out as the best trade-off between covering the complexity of all topics and not producing too many very overly specific small topics.
\ssp
\subsection{Knowledge Extraction}
\subsubsection{Citation Graph}
We extracted the list of references by identifying the keyword \code{REFERENCE} and then splitting all strings that follow by the regular expression for reference number \code{[d+]}.
Within each extracted reference, the title was separated with a \code{“(.\*)”} regular expression. These extracted reference titles were then matched against the extracted paper titles using the RapidFuzz library \cite{rapidfuzz} which allows inaccurate letter matching. Since we wanted to match potentially incomplete sub-string versions of the titles between references and papers, we utilized the token sort ratio metric which matches sub-sequences since the title is only a part of the reference.
\ssp
\section{Conclusion}

This work demonstrates the value of advanced NLP techniques for extracting insights from scientific literature, using accelerator research as a case study. We performed an analysis of IPAC and ICALEPCS conference proceedings leveraging semantic search, topic modeling, and graph methods to interpret the publication landscape solely based on the text from these publications.
We introduced a fine-tuning semantic search that enables more thorough exploration and better topic modeling.
Through unsupervised topic modeling, we uncovered latent topical structures and tracked how research priorities evolved for each conference.
Analyzing citation and bipartite graphs provided additional perspective into relationships and influences between publications, highlighting potentially impactful contributions.
We publish all data and code to empower others in the accelerator community to conduct their analyses and gain new insights from this literature.

Code and materials are available at~\href{https://github.com/sulcantonin/TEXT\_ICALEPCS23}{https://github.com/sulcantonin/TEXT\_ICALEPCS23}

\ssp
\section{Acknowledgements}

We acknowledge DESY (Hamburg, Germany), a member of the Helmholtz Association HGF, for its support in providing resources and infrastructure. Furthermore, we would like to thank all colleagues of the MCS  group for their contributions to this work and help in preparing this paper.

% \clearpage
\ifboolexpr{bool{jacowbiblatex}}%
	{\printbibliography}%
	
\end{document}